\documentclass[final,3p,times,twocolumn]{elsarticle}

\usepackage{amssymb}
\usepackage{amsthm}

\usepackage{lineno}

\usepackage{framed,multirow}
\usepackage{booktabs}

\usepackage{latexsym}

\usepackage{url}
\usepackage{xcolor}
\definecolor{newcolor}{rgb}{.8,.349,.1}

\usepackage[utf8]{inputenc}
\usepackage[small]{caption}
\usepackage{graphicx}
\usepackage{amsmath}
\usepackage{booktabs}
\usepackage[normalem]{ulem}
\useunder{\uline}{\ul}{}
\usepackage{amssymb}
\usepackage{float}
\usepackage{balance}
\usepackage{epstopdf}
\usepackage{multirow}
\usepackage{subfigure}

\theoremstyle{problem}
\newtheorem{problem}{Problem}
\theoremstyle{definition}
\newtheorem{definition}{Definition}
\theoremstyle{example}
\newtheorem{example}{Example}
\usepackage{epstopdf}
\usepackage{xcolor}
\newcommand{\advice}[1]{\textcolor{black}{#1}}

\usepackage{makecell}
\usepackage{longtable}

\usepackage{microtype}
\usepackage[noend]{algpseudocode}
\usepackage{algorithmicx,algorithm}

\journal{Knowledge-Based Systems}

\begin{document}

\begin{frontmatter}



\title{Bi-CLKT: Bi-Graph Contrastive Learning based Knowledge Tracing}


\author[1]{Xiangyu Song}
\author[1]{Jianxin Li\corref{cor1}}
\cortext[cor1]{Corresponding author.}
\ead{jianxin.li@deakin.edu.au}
\author[2]{Qi Lei}
\author[3]{Wei Zhao}
\author[4]{Yunliang Chen\corref{cor1}}

\author[5]{Ajmal Mian}

\address[1]{School of IT, Faculty of Science, Engineering and Built Environment, Deakin University, Geelong, VIC 3220, Australia}
\address[2]{School of Information Engineering, Chang’an University, Xi’an, China}
\address[3]{School of Computer Science and Technology, Xidian University, Xi’an, China}
\address[4]{School of Computer Science, China University of Geosciences, Wuhan, China}
\address[5]{Department of Computer Science and Software Engineering, The University of Western Australia, Perth, Australia}

\begin{abstract}
The goal of Knowledge Tracing (KT) is to 
estimate how well students have mastered a concept based on their historical learning of related exercises. 
The benefit of knowledge tracing is that students' learning plans can be better organised and adjusted, and interventions can be made when necessary. 
With the recent rise of deep learning, Deep Knowledge Tracing (DKT) has utilised Recurrent Neural Networks (RNNs) to accomplish this task with some success. 
Other works have attempted to introduce Graph Neural Networks (GNNs) and redefine the task 
accordingly to achieve significant improvements. However, these efforts 
suffer from at least one of the following drawbacks: 1) they pay too much attention to details of the nodes rather than to high-level semantic information; 2) 
they struggle to effectively establish spatial associations and complex structures of the nodes; and 
%
%
3) they represent either concepts or exercises only, without integrating them. 
Inspired by recent advances in self-supervised learning, we propose a \textbf{Bi}-Graph \textbf{C}ontrastive \textbf{L}earning based \textbf{K}nowledge \textbf{T}racing  
(\textbf{Bi-CLKT}) to address these limitations. Specifically, we design a two-layer comparative learning scheme based on an \advice{``exercise-to-exercise'' (E2E)} relational subgraph. \advice{It involves node-level contrastive learning of subgraphs to obtain  
discriminative representations of exercises, and graph-level contrastive learning to obtain 
discriminative representations of concepts.} Moreover, we designed a joint 
contrastive loss to obtain better representations and hence better prediction performance. Also, we explored two different variants, using RNN and memory-augmented neural networks as the prediction layer for comparison to obtain better representations of \advice{exercises} and concepts respectively. Extensive experiments on four real-world datasets show that the proposed Bi-CLKT and its variants outperform other baseline models.

\end{abstract} 




\end{frontmatter}

\section{Introduction}
%
%
%
%
With growing developments in online education platforms, vast amounts of online learning data is available to keep accurate and timely trace of students' learning status. 
To trace students' mastery of specific knowledge points or concepts, a fundamental task called Knowledge Tracing (KT) has been proposed ~\cite{corbett1994KT}, which uses a series of student interactions with exercises to predict their mastery of the concepts corresponding to those exercises. Specifically, Knowledge Tracing addresses the problem of predicting whether students will be able to correctly answer the next exercise 
relevant to a concept, given their previous learning interactions. In recent years, KT tasks have received \advice{significant attention} in academic areas, and many scholars have conducted research to propose numerous methods that deal with this problem. Conventional approaches in this domain are mainly divided into the Bayesian Knowledge Tracing model using Hidden Markov Models ~\cite{corbett1994KT} and Deep 
Knowledge Tracing using Deep Neural Networks ~\cite{ piech2015deep} and its derivative methods ~\cite{graves2014neural, zhang2017dynamic, pandey2019self, su2018exercise}. 
  

Existing KT methods ~\cite{d2008more, zhang2017dynamic, piech2015deep} generally target the concept to which the exercise belongs rather than distinguishing between the exercises themselves to build predictive models. Such an approach assumes that the ability of a student to solve the relevant exercise correctly to a certain extent directly reflects that student's mastery of the concept. Therefore, prediction based on concepts in such a way is a \advice{viable option}, however, this reduces the difficulty of the task itself, given the limited performance of the model. Generally, a KT task comprises multiple concepts and a large number of exercises with an even larger number of situations where a concept is associated with many exercises, and a proportion of situations where an exercise may correspond to multiple concepts. Traditional models can only deal with the former, while for the latter, they often have to resort to \advice{dividing these cross-concept exercises} into multiple single-concept exercises. Such an approach, while enhancing the feasibility of these models, nevertheless interferes with the accuracy of the overall task.

%
Although these concept-based KT methods have been somewhat successful, the characteristics of the exercises themselves are often overlooked. This can lead to a reduction in the ultimate predictive accuracy of the model and a failure to predict specific exercises. Even if two exercises have the same concept, the difference in their difficulty level may ultimately lead to a large difference in the probability of them being answered correctly. Therefore, some previous works~\cite{minn2019dynamic, dos2016multilabel, Yang2020GIKTAG, abdelrahman2019knowledge,xue2021dynamic, song2020sepn} have attempted to use exercise features as a supplement to concept input, achieving success to some extent. However, due to the relatively large difference between the number of exercises and the number of \advice{exercises} students actually interact with, each student may only interact with a very small fraction of the exercises, leading to problems of sparse data. Furthermore, for those exercises that span concepts, simply adding features to the exercises loses potential inter-exercise and inter-concept information. Therefore, the use of higher-order information such as ``exercise-to-exercise'' (E2E) and ``concept-to-concept'' (C2C) is necessary to address these issues.


\advice{The idea of redefining the Knowledge Tracing problem in terms of graphs has recently gained significant momentum due to the widespread deployment of GNNs ~\cite{nakagawa2019graph, Yang2020GIKTAG, liu2020improving, li2021deep, yin2021deep} and breakthroughs in addressing the \advice{unpredictability} of traditional approaches to cross-concept exercises.}
%
%
Traditional KT usually takes sequential data as input in the form of concepts corresponding to the input exercises and their responses. This leads to a lack of information between exercises, and only the relationship between exercises and concepts is available. Recent research in graph theory has opened up the possibility of breaking this bottleneck. Unlike sequential data, graph data can capture the higher order information of ``exercise-to-exercise'' (E2E) and ``concept-to-concept'' (C2C) very well due to the multivariate node and edge structure of the graph itself. As a result, some research~\cite{nakagawa2019graph, song2021jkt} has turned to redefining this task in terms of graphs. \advice{However, these efforts face several problems: 1) too much attention to the details of the nodes rather than high-level semantic information; 2) difficulty in effectively establishing spatial associations and complex structures of the nodes; and 3) representing only concepts or exercises without integrating them. }

%
%
Due to the difficulty and inaccuracy of data labelling, self-supervised learning has become increasingly popular, with great success in many areas such as computer vision ~\cite{ bachman2019learning, he2020momentum, tian2020contrastive} and natural language processing ~\cite{collobert2008unified, mnih2013learning}. The speciality of self-supervised learning is in dealing with low quality or missing labels, which is a requirement for supervised learning and uses the input data itself for incremental layers as supervised labels for the learning model. \advice{This can be as powerful as supervised models with \advice{specifically} labelled information and eliminates the tedious labelling work required by supervised models.} Specifically, self-supervised learning eliminates the need to label specific tasks which is the biggest bottleneck in supervised learning. Especially for large amounts of network data, obtaining high quality labels at scale is often very expensive and time consuming. Self-supervised learning has been shown to excel in tasks with textual and image datasets, but is still in its infancy for problems on the graphs such as retrieval, recommendation, graph mining, and social network analysis.

\advice{In this paper, we address the problems encountered by traditional GNN based KT models and propose a self-supervised learning framework along with Bi-Graph Contrastive Learning based Knowledge Tracing (Bi-CLKT) model. Our model employs contrastive learning with global-bilayer and local-bilayer structures, where they apply graph-level and node-level GCNs to extract ``exercise-to-exercise'' (E2E) and ``concept-to-concept'' (C2C) relational information, respectively. Finally, the prediction on students' performance is carried out by a prediction layer based on a deep neural network.}

Our approach has been fully validated using four benchmark datasets. In terms of prediction performance, our model and its variants outperformed traditional deep learning-based methods, showing great potential for knowledge-tracing prediction accuracy. Furthermore, through a series of ablation studies, we have analysed each module separately, making the model more interpretable. Our proposed model also has some technical innovations and improvements. Due to the specificity of this task, which requires separate representation learning for two related but independent entities, our two-layer comparative learning structure fits well into this. Our specific contributions are described as follows: 
\begin{itemize}
    \item To the best of our knowledge, we present the first self-supervised learning based knowledge-tracing framework. Through contrastive self-supervised learning, we solve a number of problems encountered by traditional GNN-based knowledge-tracing models leading to significant improvement in the accuracy of the final prediction results.
    \item For Knowledge Tracing, we design a two-layer contrastive learning framework, which performs ``exercise-to-exercise'' (E2E) and ``concept-to-concept'' (C2C) relational information at the global and local levels respectively. The representation of exercise is learned eventually and effectively combined by a joint contrastive loss function. Such a structure allows an exercise embedding to have both exercise and concept structural information, which has a positive effect on the final prediction task.
    \item We perform thorough experiments on four real-world open datasets, and the results show that our proposed framework and its variants all have a significant improvement in predictive efficiency compared to the individual baseline models. We also perform
    ablations studies to analyse the validity of each individual module, which greatly enhances the model interpretability.
\end{itemize}

%
%


\section{Related Work}
\subsection{Knowledge Tracing}

There are two main approaches to using machine learning for Knowledge Tracing. The first one is the traditional machine learning KT approach represented by Bayesian Knowledge Tracing (BKT) ~\cite{corbett1994KT}.  
%
%
BKT primarily applies the Hidden Markov Model, which uses Bayesian rules to update the state of each concept considered as a binary variable.
Several works have extended the basic BKT model and introduced additional variables such as slip and guess probabilities~\cite{d2008more}, concept difficulty ~\cite{pardos2011kt} and \advice{student personalisation~\cite{pardos2010modeling, yudelson2013individualized,song2020sepn}}. On the other hand, traditional machine learning KT models also include factor analysis models such as item response theory (IRT)~\cite{ebbinghaus2013memory} and performance factor analysis (PFA) ~\cite{pavlik2009performance,li2021deep}, which tend to focus on learning general parameters from historical data to make predictions.

%
With the development of Deep Neural Networks, the literature has experienced advances in Deep Knowledge Tracing methods that have proved to be more effective in learning valid representations for large amounts of data for more accurate predictions. 
For example, Deep Knowledge Tracing (DKT) ~\cite{piech2015deep} uses recurrent neural networks (RNNs) to track students' knowledge states and became the first deep KT method to achieve excellent results. Another example is the Dynamic Key-Value Memory Network (DKVMN) ~\cite{zhang2017dynamic} that builds a static and dynamic matrix to store and update all concepts and students' learning states respectively. Xu et al.~\cite{xu2018deep} propose a pioneering deep matrix factorization method for conceptual representation learning from multi-view data. However, these classical models consider the most basic concept features only, and the absence of exercise features leads to 
%
%
unreliable final predictions.

Some deeper KT methods have since been proposed that do take into account the features of the exercises in their predictions. For example, Exercise-Enhanced Recurrent Neural Network with \advice{attention mechanism (EERNNA)}~\cite{su2018exercise} uses information about the text of the exercise to allow the embedding itself to contain the features of the exercise, but in reality it is difficult to collect such textual information, and doing so introduces too much interference into the embedding itself. Dynamic Student Classification on Memory Networks (DSCMIN)~\cite{minn2019dynamic} uses modelling of problem difficulty to help distinguish between different problems under the same concept. DHKT, on the other hand, augments DKT by using relationships between problems and skills to obtain a representation of the exercises. However, this does not capture the relationships between exercises and concepts due to the data sparsity issues. 
Due to the presence of long-term dependencies in practice sequences, the Sequence Key Value Memory Network (SKVMN)~\cite{abdelrahman2019knowledge} has improved the LSTM with good results in order to improve the ability to capture such dependencies. Our approach differs from these methods in that they build the graph of exercise-influence relations from the original \advice{``exercise-to-concept'' (E2C)} relations by certain assumptions and use graph-level and node-level GCNs, respectively, to extract the ``exercise-to exercise'' (E2E) and ``concept-to-concept'' (C2C) relational information. On the other hand, to reduce the interference of too much detailed information, we use contrast learning model to learn the concepts and exercises separately for representation.

\subsection{Self-supervised Learning}
Research on self-supervised learning can be broadly divided into two branches: generative models and contrastive models. The main representative of the generative model is the automatic coding which is very popular. The main approach on graph data is to learn the embedding of nodes of the graph into a latent space through GNN, and then reconstruct the structure and properties of the original graph through the learned representations. The representations of the nodes are adjusted by reducing the size of the loss between \advice{the generated graph and the original graph} step by step. The learned representations are then used to reconstruct the original diagram. These representations encode the structural and attribute features of the original graph.
\advice{Contrastive} learning, on the other hand, uses augmentation methods to structurally disrupt the input data, separating out the predicted objects and corresponding labels from their own structure before learning the representations, and finally comparing the loss functions to minimize the distance between positive pairs and maximize the distance between negative pairs to achieve a structural grasp of the complete graph. Pioneer methods along the direction of learning graph representations of GCNs include Hu et al. ~\cite{kingma2014adam} and Kaveh et al. ~\cite{li2017attributed}.

\subsection{Contrastive Learning on Graphs}

Contrastive learning is a type of self-supervised learning where the target label to be learned is generated from source data itself. It brings similar representations closer together and dissimilar ones further apart by comparison. For graph data, traditional learning methods~\cite{becker1992self, wu2018unsupervised, ye2019unsupervised} often overemphasise detailed information at the expense of structural information. On the other hand, contrastive learning compensates for this by nicely finding a balance between local and global representation learning. Although contrastive learning on graph data is still in its infancy, it has been demonstrated by numerous models to be powerful in its control of graph structural information.


\advice{\section{Preliminary and Problem Statement}}

In this section, we define the task of Knowledge Tracing in our setting. We first formally define the student performance prediction problem, which represents the level of student mastery of each concept by the accuracy with which students interact with the exercises under each concept. Next, we present the important definitions used in our study.



\advice{\subsection{Problem Definition}}
In the Knowledge Tracing task, we record the a particular student's practice process as $s\in\mathcal{S}$, $s=\left \{\mathcal{X}_{0}...\mathcal{X}_{T}\right \}=\left \{(e_{0}, a_{0}),(e_{1}, a_{1}),...,(e_{T}, a_{T}) \right \},$ where $|S|$ represents the student, and $|E|$ represents the specific exercise, $e_{t} \in E$ represents the exercise that student $s$ does in its exercise step $t$, and $r_{t}$ represents the correctness or otherwise of the corresponding exercise. In general, $r_{t}$ equals 1 if the student $s$ answered the exercise $e_{t}$ correctly, otherwise $r_{t}$ equals 0. To trace the mastery of a specific concept $c\in\mathcal{C}$, we normally observe a sequence of a student's interactions $e_{t} \in E$ and predict the result $r_{t+1}$ of the next exercise $e_{t+1}$. Finally, the probability of a student getting any next exercise correct for a specific concept is taken as the student's mastery of that concept. The specific definitions are as follows:

\begin{problem}[{Performance Prediction in KT}]
Give a sequence of interaction observations taken by a students $s\in\mathcal{S}$, $s=\left \{\mathcal{X}_{0}...\mathcal{X}_{T}\right \}=\left \{(e_{0}, a_{0}),(e_{1}, a_{1}),...,(e_{T}, a_{T}) \right \}$ on a specific concept $c\in\mathcal{V}_{c}$, where $\mathcal{X}_{t} = \left \{e_{t}, a_{t} \right \}$ for the exercise $e\in\mathcal{V}_{e}$ being answered at the time step $t\in T$ with whether or not the exercise was answered correctly $a_{t}\in\left \{0,1 \right\} $. The objective of Knowledge Tracing task is to predict the next interaction \advice{$\mathcal{X}_{t+1}$}.
\end{problem}

To solve this problem, we dig deeper into the ``exercise-to-exercise"(E2E) graph structure relationships from the original ``exercise-to-concept" sequence data. The secondary ``exercise-to-exercise"(E2E) relationships extracted are used to construct an exercise influence graph based on each of the different concepts. Then, we obtained separate pre-representations of the exercises and concepts by applying graph-level and node-level GCNs on these exercise influence graphs. Finally, we fuse them into a contrast learning-based Knowledge Tracing graph to train these representations to achieve optimal results under the Knowledge Tracing task.

The exercise-level influence graph is derived from students' transitions between exercises. It assumes that the majority of students who get two different exercises correct at the same time will have a high degree of similarity or correlation between the two exercises, and therefore will have relatively high weights between them, and vice versa. The creation of the exercise-level influence graph not only ameliorates the 
existing models' shortcoming of not being able to distinguish between exercises under the same concept, but also provides rich information on the structure of the \emph{``exercise-to-exercise"} graph.
\begin{figure}[h]
\centering
{\includegraphics[scale=0.3]{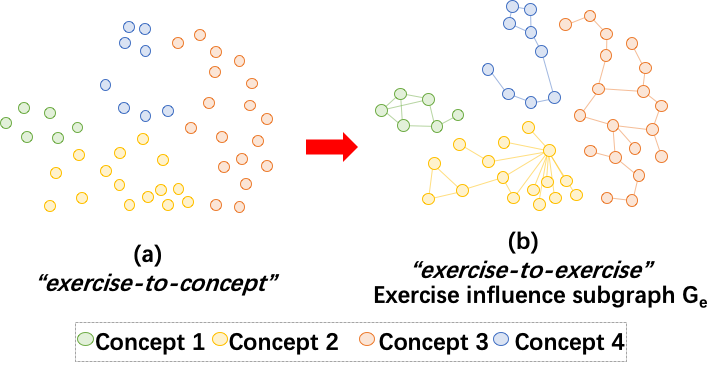}}
\caption{An example of creating an exercise influence subgraph from an exercise-to-concept relationship graph.} \label{figure1}
\end{figure}
\begin{definition}(Exercise-level influence graph) 
Given an exercise set $\mathcal{V}_{e}$ and an ``exercise-to-exercise" interaction set $\mathcal{E}_{e}$, an exercise influence sub-graph is a graph $\mathcal{G}_{e}=(\mathcal{V}_{e}, \mathcal{E}_{e})$ that takes the exercises in $\mathcal{V}_{e}$ as the vertices and the interactions in $\mathcal{E}_{e}$ as the edges. The weighted influence $Q_{e}^{ij}$ $(i,j\in \mathbb{N},i,j\leqslant m)$ of the edges of co-ocurred exercises $\mathcal{V}_{e}^{i}$ and $\mathcal{V}_{e}^{j}$ is measured by 
\begin{equation}
Q_{e}^{ij}=\frac{f^{c}(\mathcal{V}_{e}^{i}, \mathcal{V}_{e}^{j})}{\sum_{m}f^{o}(\mathcal{V}_{e}^{i}, \mathcal{V}_{e}^{m})}
\end{equation}
where $Q_{e}^{ij}$ is the co-correctness rate of exercises $\mathcal{V}_{e}^{j}$ and $\mathcal{V}_{e}^{i}$ among all the answered co-occurred exercises involving \advice{$\mathcal{V}_{e}^{i}$}. $f^{c}$ and $f^{o}$ denote the count of co-correctness and co-occurrence respectively. Edge $\mathcal{E}_{e}$ ($\mathcal{V}_{e}^{j}$, $\mathcal{V}_{e}^{i}$) only exists when $Q_{e}^{ij}$ is larger than a certain threshold $\tau$.
\end{definition}

\begin{example} As shown in Figure \ref{figure1}, we constructed an example of an exercise influence subgraph by assuming a high degree of similarity in the questions that a student can answer correctly at the same time, in which the properties of different exercises are reflected from node to node. In turn, the properties of different concepts are reflected between different subgraphs. They represent ``exercise-to-exercise'' (E2E) and ``concept-to-concept'' (C2C) relational information, respectively.
\end{example}


\advice{\subsection{Representative Solutions Study}}
\advice{The vector sequence of $\mathbf{x}_{1}, \ldots, \mathbf{x}_{T}$ as input to conventional deep knowledge tracking is mapped to the output vector sequence $\mathbf{y}_{1}, \ldots, \mathbf{y}_{T}$ by computing a sequence of 'hidden' states $\mathbf{h}_{1}, \ldots, \mathbf{h}_{T}$. This can be seen as a continuous encoding of information about \advice{historical learning performance} to make predictions about the future, and DKT makes the connection between input and output via a simple recurrent neural network. The input $\left(\mathbf{x}_{t}\right)$ to the dynamic network is a representation of the student's historical behaviour, while the prediction $\left(\mathbf{y}_{t}\right)$ is a vector representing the probability of being correct for each sample exercise. More details are defined by the equations:}

\advice{
\begin{equation}
\begin{aligned}
&\mathbf{h}_{t}=\tanh \left(\mathbf{W}_{h x} \mathbf{x}_{t}+\mathbf{W}_{h h} \mathbf{h}_{t-1}+\mathbf{b}_{h}\right) \\
&\mathbf{y}_{t}=\sigma\left(\mathbf{W}_{y h} \mathbf{h}_{t}+\mathbf{b}_{y}\right)
\end{aligned}
\end{equation}
}

\advice{where the sigmoid and tanh functions $\sigma(\cdot)$ are used as activation functions, $\mathbf{W}_{h h}$ denotes the input weight matrix, $\mathrm{h}_{0}$ denotes the initial state, \advice{$\mathrm{W}_{y h}$} denotes the readout weight matrix and $\mathbf{W}_{h h}$ denotes the cyclic weight matrix. The deviations of the latent and readout cells are given by $\mathrm{b}_{h}$ and $\mathrm{b}_{y}$.}



\advice{Another classical approach is DKVMN, which outputs the probability of a response $p\left(r_{\mathrm{t}} \mid q_{\mathrm{t}}\right)$ through a discrete exercise label $q_{t}$. The motion and response tuples $\left(q_{t}, r_{t}\right)$ are then updated. Here, $q_{t}$ is a set with $Q$ distinct exercise labels and $r_{t}$ is the binary value of whether the student got it right or not. DKVMN assumes that the exercise is based on the set of potential concepts $\left\{c^{1}, c^{2}, \ldots, c^{N}\right\}$ with $N$. The key matrix $\mathrm{M}^{k}$ (size $N \times d_{k}$) is used to store these concepts of size $N\times d_{k}$. Concept states $\left\{\mathbf{s}_{t}^{1}, \mathbf{s}_{t}^{2}, \ldots, \mathbf{s}_{t}^{N}\right\}$ are stored as students' mastery of each concept in the time-varying value matrix $\mathbf{M}_{\mathrm{t}}^{\mathrm{v}}$ (size $N \times d_{\mathrm{v}}$). Ultimately, DKVMN tracks student knowledge by reading and writing to the value matrix using the relevant weights computed from the input exercises and the key matrix.}

In the KT process, existing models often do not link the different concepts well, which leads to the inability of these models to make correct or complete predictions when students encounter exercises on concepts that have not been covered before or when an exercise involves multiple concepts. We, therefore, fill this gap by using the construction of exercise influence subgraphs, with nodes connecting different exercises to each other and subgraphs connecting different concepts to each other.

\advice{By building an exercise influence subgraph, we can transform the original sequence data into graph structured data containing ``exercise-to-exercise'' (E2E) and ``concept-to-concept'' (C2C) relational information. We then learn these data into the corresponding embedded representations by using node-level and graph-level GCNs respectively, while learning the best architecture by contrast to maximise the differentiation of these representations so that the final representations contain both information about the exercises and concepts, and their differentiation from other concepts and exercises.}




\section{The Bi-Graph Contrastive Knowledge Tracing}
\advice{Inspired by recent developments of contrastive learning in visual representation, an increasing scale of research has shown that contrastive learning frameworks perform well on graph-structured data as well. Therefore, after various comparisons and studies, we propose a Bi-graph Contrastive Knowledge Tracing representation learning (Bi-CLKT) based on graph-level (global) and node-level (local) GCNs. In the next section, we describe Bi-CLKT in detail, first by briefly discuss a traditional contrastive learning framework, and then more specifically by presenting our proposed Bi-graph contrastive learning framework. Finally, we provide the theoretical rationale behind our approach. }

\advice{\subsection{The Graph CL Paradigm} }
\begin{figure*}[h]
    \centering
    \includegraphics[scale=0.45]{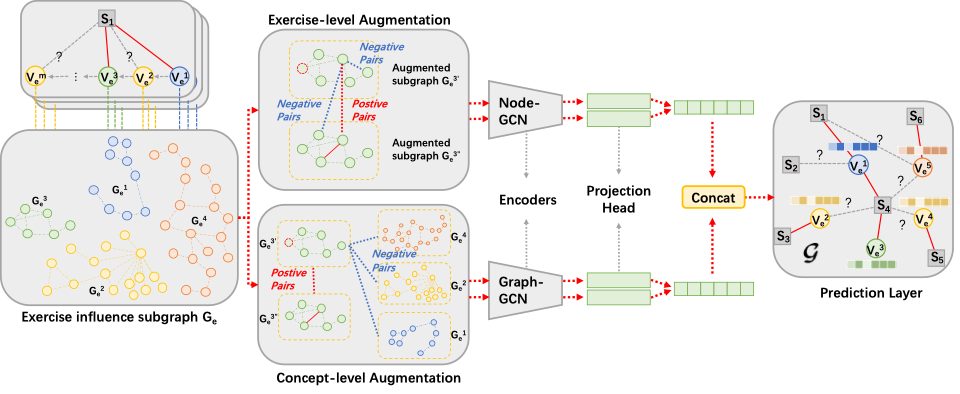}
    \caption{An overview of Bi-Graph Contrastive Knowledge Tracing framework.}
    \label{fig:DKT_figure3}
\end{figure*}
As shown in Figure 2, our proposed Bi-CLKT framework extends the common graph CL paradigm. Common graph CL paradigms typically employ either global features or local features to seek to maximise the consistency of representations between the graph-level or node-level of different views. Specifically, graph CL paradigms first generate two graph views by performing random graph \advice{augmentations} on the original graph, such as eliminating or adding edges, eliminating nodes, masking attributes, etc. For the global graphCL, these two views are treated as a positive pair, while the other graphs are treated as negative pairs. \advice{While} for the local graphCL, the nodes in these two views are found as a positive pair and negative pairs. we then employ a contrastive loss that forces the positive pair's embeddings in the view are consistent with each other, while trying to distance all negative pairs from each other. Specifically, the graph CL paradigm consists of four main components:

\begin{itemize}
    \item Graph data augmentation. The given graph $\mathcal{G}$ is augmented by eliminating or adding edges, eliminating nodes, masking attributes, etc. to obtain two related views $\hat{\mathcal{G}}_{i}, \hat{\mathcal{G}}_{j}$ (i.e. augmented graphs), which are treated as a positive pair in the global graph CL. Whereas in the local graph CL, all \advice{positive} pairs and negative pairs exist in the both two views.
    \item GCN-based encoder. GCN-based encoders node-level $\boldsymbol{f}_{\phi}$ and graph-level $\boldsymbol{g}_{\theta}$ for augmented graphs $\hat{\mathcal{G}}_{i},\hat{\mathcal{G}}_{j}$ to extract representation vectors $\boldsymbol{h}_{j}$.
    \item Projection head. A non-linear transformation $h(\cdot)$ named projection head maps augmented representations to another latent space where the contrastive loss is calculated. In graph contrastive learning, $\boldsymbol{z}_{i}, \boldsymbol{z}_{j}$ is obtained by a Multi-Layer perceptron (MLP).
    \item Joint contrastive loss function. The joint contrastive loss function $\mathcal{L}(\cdot)$ is defined as forcing the maximisation of the distance between the positive pair $\boldsymbol{z}_{i}, \boldsymbol{z}_{j}$ and the negative pair in the two \advice{subgraphs respectively}. The final normalized temperature-scaled cross entropy loss is used to calculate the loss of the two contrastive learning modules.

\end{itemize}

\advice{\subsection{The Bi-CLKT Framework}}

In general, common graph CL approaches usually choose either global graph CL at the graph level or local graph CL at the node level, which seek to learn representations by maximising the consistency between views from different perspectives. Normally, these approaches often have only one feature, global or local. In the Bi-CLKT model, due to the specificity of the KT task, we need to acquire both ``exercise-to-exercise'' (E2E) and ``concept-to-concept'' (C2C) relational information. 
These two relational features respectively have the corresponding properties of being local and global. Therefore, we propose to design a Bi-graph contrastive learning framework with both local and globe features.

\advice{To match this two-layer framework, we use node-level GCN to learn ``exercise-to-exercise'' (E2E) embedding and graph-level GCN to learn ``concept-to-concept'' (C2C) embedding, respectively. In the process of graph data augmentation, we design separate graph augmentation processes for each of these two layers of the contrastive learning framework. We mainly augment the input graph by randomly removing edges and masking node features in the graph. In addition, inspired by~\cite{zhu2021graph}, we introduce differences in the importance of different nodes and edges by calculating the centrality of different nodes through methods such as random walk and PageRank, and those nodes and edges with lower importance are prioritized for elimination during the graph augmentation process.}

\advice{\subsubsection{Graph Data Augmentation}}
For the exercise level augmentation, we took two different approaches to augmenting the input graph, i.e. randomly picking edges or points to be removed. Formally, we form a modified subset $\widetilde{\mathcal{E}} $ and $\widetilde{\mathcal{V}}_{e}$ by randomly selecting some nodes and edges from the original $\mathcal{E}$ and $\mathcal{V}_{e}$ in proportion to the probability of random selection of \begin{equation}
\begin{array}{l}
P_{E}\{(u, v) \in \widetilde{\mathcal{E}}\}=1-p_{u v}^{e}\\
P_{V_e}\{v \in \widetilde{\mathcal{V}}_{e}\}=1-p_{v}^{e}
\end{array}
\end{equation} 
%
%
where $p_{u v}^{e}$ and $p_{v}^{e}$ are the probabilities of eliminating edges $(u, v)$ and nodes $v$, and then $\widetilde{\mathcal{E}}$ and $\widetilde{\mathcal{V}}_{e}\}$ are the set of edges and the set of shop points after graph augmentation. The $p_{u v}^{e}$ and $p_{v}^{e}$ reflect the importance of edges $(u, v)$ and nodes $v$, respectively. By doing so, this function ensures that the unimportant edges or points are eliminated preferentially while ensuring that the important structure of the graph is not compromised.

In concept-level graph augmentation, similar to the increments at the exercise level, we need to compute probabilities $p_{u v}^{e}$ and $p_{v}^{e}$ to reflect the importance of the corresponding edges $(u, v)$ and nodes $v$. The difference is that in concept-level graph augmentation, to better fit the graph-level GCN, we use the PageRank algorithm to combine $p_{u v}^{e}$ and $p_{v}^{e}$ to work out a composite probability $p_{i}^{f}$ for the concept level. The specific calculation is defined below as defined below
\begin{equation}
p_{i}^{f}=\min \left(\frac{s_{\max }^{f}-s_{i}^{f}}{s_{\max }^{f}-\mu_{s}^{f}} \cdot p_{f}, p_{\tau}\right)
\end{equation}
where we compute the maximum and average values of $s_{i}^{f}$ and denote them by $s_{i}^{f}=\log w_{i}^{f}, s_{max }^{f}$ and $mu_{s}^{f}$ respectively, and $p_{f}$ is the probability of combining $p_{u v}^{e}$ and $p_{v}^{e}$.

Finally, we generate two corrupted graph views $\widetilde{\mathcal{G}}_{1}, \widetilde{\mathcal{G}}_{2}$ by augmentation for the exercise and conceptual levels, respectively. In Bi-CLKT, the probabilities of generating the two views are different, and to better target the description of these two probabilities we denote them by $p_{f,1}$ and $p_{f,2}$ respectively.

\advice{\subsubsection{Node-Level and Graph-Level Encoder}}
\advice{
Direct use of the node features of the last layer of the encoder $K$ is the most direct way to obtain a node-level representation $h_{v}$ of node $v$, i.e. $h_{v}=x_{v}^{(K)}$. Where the use of skip connections or skip knowledge to generate node-level representations is do commonly used. However, the method of connecting the node features of all layers produces a node-level representation with a different dimensionality than the node features. To avoid this problem, we perform a linear transformation of the node features of all layers before joining them together.
\begin{equation}
h_{v}=\operatorname{CONCAT}\left(\left[x_{v}^{(k)}\right]_{k=1}^{K}\right) \boldsymbol{W},
\end{equation}
where the weight matrix that reduces the size of the $h_{v}$ dimension is $\boldsymbol{W} \in \mathbb{R}^{\left(\sum_{k} d_{k}\right)\times d}$.}


.

\advice{The key operation for computing the node-level representation of a graph $h_{\text {graph }}$ is the READOUT function, of which summation and averaging are the most commonly used READOUT functions. For reasons of node envelope invariance. We use a summation over all node representations, i.e.
\begin{equation}
h_{graph}=\operatorname{READOUT}(H)=\sigma\left(\sum_{v=1}^{|V|} h_{v}\right),
\end{equation}
where $\sigma$ is the sigmoid function and $|V|$ denotes the total number of nodes in a given graph.}

Due to the nature of Bi-CLKT's two-layer structure, we have used node-level and graph-level GCNs as encoders for the exercises and concepts respectively. The specific structural form is as follows   
\begin{equation}
\begin{aligned}
\mathrm{GC}_{i}(X, A) &=\sigma\left(\hat{D}^{-\frac{1}{2}} \hat{A} \hat{D}^{-\frac{1}{2}} X W_{i}\right) \\
f(X, A) &=\mathrm{GC}_{2}\left(\mathrm{GC}_{1}(X, A), A\right)
\end{aligned}
\end{equation}

where the adjacency matrix is $\hat{A}=A+I$ and the degree matrix is $\hat{D}=$ $\sum_{i} \hat{A}_{i}$ and is the non-linear activation function we use $\sigma(\cdot)$.



\advice{\subsubsection{Projection Head}}
Furthermore, in Bi-CLKT we map the enhanced representations to a uniform latent space via a non-linear transformation called projection head $g(\cdot)$, where the contrast loss is computed. In graphical contrast learning, a multilayer perceptron (MLP) is applied to obtain these mappings $z_{i}, \bar{z}_{j}$.


\advice{In terms of projection head, the case of considering view generation in terms of mutual information is that a good view generation should minimise the MI between two views $I\left(v_{1}, v_{2}\right)$, provided that $I\left(v_{1}, y\right)=I\left(v_{2}, y\right)=I(x, y)$. Intuitively, the fact that the generated viewpoints $v_{i}$ do not affect the information that determines the prediction of the downstream task can guarantee the effectiveness of contrast learning. Thus under this restriction, divergence between viewpoints as it increases leads to better learning results. From a flowform perspective, we adopt the extension hypothesis and find that an increase in data can induce continuity in the neighbourhood of each instance.}

\subsection{Joint Contrastive Loss Function}

To better learn the representations of exercises and concepts, we used a joint contrastive loss function $\mathcal{L}(\cdot)$. This loss function forces the maximisation of the distance between positive pairs of previously learned mappings $z_{i}, z_{j}$ and other negative pairs. With extensive comparisons, we found that normalised temperature scaled cross-entropy loss (NT-Xent) was the most appropriate loss model. In the node-level GCN training process, we randomly draw N small batches of nodes on the practice influence graph under the same concept, and learn these nodes by local contrast learning, where all 1-hop neighbour points under its same view are used as his negative pairs, along with all 1- hop neighbours of the other view, and the only positive pairs are the corresponding points under both views. In the training process of the graph-level GCN, we randomly draw $N$ small batches of exercise influence graphs, so that $2 N$ augmented graphs are generated as positive pairs, while all other graphs are used as their negative pairs.
The cosine similarity function is denoted as 
\begin{equation}
    \operatorname{sim}\left(z_{n, i}, \boldsymbol{z}_{n, j}\right)=\boldsymbol{z}_{n, i}^{\top} z_{n, j} /\left\|z_{n, i}\right\|\left\|z_{n, j}\right\|
\end{equation}

\advice{As for the objective function, the prevailing practice is to use a standard binary cross-entropy loss between positive and negative examples i.e. a noisy contrast type objective. However, we have found by research that it is detrimental to representation learning if positive and negative examples are absolutely distinguished. This is mainly due to the fact that these contextual subgraphs are extracted from the same original graph and overlap each other. Therefore, we used normalised temperature scaled cross-entropy loss (NT-Xent) for model optimisation so that positive and negative samples are well differentiated to some extent, resulting in a high quality representation.} The NT-Xent of the $n$-th graph is defined as
\begin{equation}.
\ell_{n}=-\log \frac{\exp \left(\operatorname{sim}\left(z_{n, i}, z_{n, j}\right) / \tau\right)}{\sum_{n^{\prime}=1, n^{\prime} \neq n}^{N} \exp \left(\operatorname{sim}\left(z_{n, i}, z_{n^{\prime}, j}\right) / \tau\right)}
\end{equation}
where the temperature parameter is denoted by $\tau$. These last two correspond to the exercise and concept losses being computed in all \advice{positive pairs}, respectively.


\section{Experimental Settings and Results}

We performe extensive experiments on four real-world datasets to evaluate the performance of our Bi-CLKT model. We also compare it to several state-of-the-art machine learning and deep learning Knowledge Tracing models. To fully evaluate the performance of Bi-CLKT, we conducted a large number of in-depth ablation experiments, which validated the role of each module in Bi-CLKT and enhanced the interpretability of the model.

\advice{To implement the baseline and Bi-CLKT models, we used PyTorch and the Geometric Deep Learning extension library. Experiments were conducted on four NVIDIA TITAN V GPUs. Bi-CLKT was used to learn node representations in a self-supervised contrastive learning fashion, and these representations were then used to evaluate node-level and graph-level classifications. This was done by directly using these representations to train and test a simple linear (logistic regression) classifier. In pre-processing, we perform line normalisation of and apply processing strategies. We normalise the learned embeddings before feeding them into the logistic regression classifier. In training, we use the Adam optimiser with an initial learning rate of 0.001 and the subgraph size does not exceed 20. the dimensionality of the node representation is 1024. the marginal value of the loss function is 0.75.}

\subsection{Datasets}
To evaluate our model, the experiments are conducted on the following four widely-used datasets in KT and the detailed statistics are shown in Table \ref{table1}.

\begin{itemize}
    \item \textbf{ASSISTment 2009}\footnote{https://sites.google.com/site/assistmentsdata/home/assistment-2009-2010-data} is provided by the online tutorial website ASSISTment, which is widely used to validate KT problems. Among other things, this dataset comes with accurate labels, practice and conceptual clarity. We have not modified this dataset much except for filtering out corrupt samples.
    \item \textbf{ASSISTment 2015}\footnote{https://sites.google.com/site/assistmentsdata/home/2015-assistments-concept-builder-data} is similarly from the online tutoring site ASSISTment, which further clarifies the data set ASSISTment collected in 2009 by collapsing the number of concepts to exactly 100 and introducing a larger number of students, but with a slightly reduced average student interaction record.
    \item\textbf{ASSISTment Challenge}\footnote{https://sites.google.com/view/assistmentsdatamining/dataset} (ASSISTment chall) is collected for a data mining competition run by ASSISTments in 2017. It has a relatively rich average number of records per student, and because it is used for competition, the dataset as a whole has the highest degree of completeness and normality of the three datasets from ASSISTment.
    \item \textbf{STATICS 2011}\footnote{https://pslcdatashop.web.cmu.edu/DatasetInfo?datasetId=507} differs from the previous three datasets in that it is course-specific i.e. there is high relevance in the data. This dataset contains 189,297 interactions between 333 students on 1223 concepts making it the most intensive of all four datasets.
\end{itemize}

Table \ref{table1} presents all the statistical data for the dataset, where $N_S$, $N_c$ and $N_\mathcal{E}$ represent the number of students, concepts and interactions respectively.   

\begin{table}[H]
\centering
\caption{Statistics for the datasets.}
\begin{tabular}{cccc}
\hline \multirow{2}{*}{ Datasets } & \multicolumn{3}{c} { Statistics } \\
\cline { 2 - 4 } & \#$N_S$ & \#$N_c$ & \#$N_\mathcal{E}$  \\
\hline ASSISTment 2009 & 4,151 & 110 & 325,637 \\
ASSISTment 2015 & 19,917 & 100 & 708,631 \\
ASSISTment chall & 686 & 102 & 942,816  \\
STATICS 2011 & 333 & 1,223 & 189,297  \\
\hline
\end{tabular}
\label{table1}
\end{table}

\begin{table*}[ht]

\caption{Area Under the curve (AUC) and Accuracy (ACC) on four datasets. The best performing runs per metric per dataset are marked in boldface.}
\centering
\begin{tabular}{l|cc|cc|cc|cc}
\hline\hline
\multicolumn{1}{c|}{} & \multicolumn{2}{c|}{\textbf{ASSISTment 2009}} & \multicolumn{2}{c|}{\textbf{ASSISTment 2015}} & \multicolumn{2}{c|}{\textbf{ASSISTment Chall}} & \multicolumn{2}{c}{\textbf{STATICS 2011}} \\ \hline\hline
\multicolumn{1}{c|}{} & \textit{AUC}          & \textit{ACC}          & \textit{AUC}          & \textit{ACC}          & \textit{AUC}           & \textit{ACC}          & \textit{AUC}         & \textit{ACC}        \\ \hline
BKT \cite{corbett1994KT}         & 0.648                 & 0.594                 & 0.616                 & 0.592                 & 0.562                  & 0.555                 & 0.719                & 0.698               \\
DKT  \cite{piech2015deep}        & 0.74                  & 0.708                 & 0.701                 & 0.68                  & 0.691                  & 0.712                 & 0.815                & 0.723               \\
DKVMN \cite{zhang2017dynamic}       & 0.739                 & 0.618                 & 0.705                 & 0.68                  & 0.689                  & 0.614                 & 0.814                & 0.722               \\
SAKT  \cite{pandey2019self}       & 0.735                 & 0.679                 & 0.721                 & 0.647                 & 0.701                  & 0.657                 & 0.803                & 0.797               \\
EKT   \cite{liu2019ekt}       & 0.754                 & 0.702                 & 0.737                 &  0.754           & 0.72                   & 0.727                 & 0.842                &  0.819         \\
SAINT+  \cite{choi2020towards}      &  0.782           &  0.718           &  0.754           & 0.741                 &  0.734            & 0.718                 &  0.853          & 0.808               \\ \hline
 \textbf{Bi-CLKT}        & \textbf{0.857}        & \textbf{0.802}        & \textbf{0.765}        & \textbf{0.757}        & \textbf{0.775}         & \textbf{0.764}        & \textbf{0.865}       & \textbf{0.835}          \\
\hline\hline

\end{tabular}
\label{table2}
\end{table*}

\subsection{Evaluation Methods}
We compare our proposed Bi-CLKT with the following baseline methods. \begin{itemize}
    \item\textbf{Bayesian Knowledge Tracing} \cite{corbett1994KT} is a classical machine learning Knowledge Tracing model based on the Hidden Markov Model, which uses Bayesian rules to update the state of each concept, considered to be a binary variable. 
    \item\textbf{Deep Knowledge Tracing }\cite{piech2015deep} uses recurrent neural networks (RNNs) to track students' knowledge states and became the first deep KT method to achieve excellent results.
    \item\textbf{Dynamic Key-Value Memory Networks} \cite{zhang2017dynamic}, inspired by memory-enhanced neural networks, builds a static and dynamic matrix to store and update all concepts and students' learning states respectively.
    \item\textbf{SAKT} \cite{pandey2019self} is the first self-attentive based Knowledge Tracing model. \advice{It abandons the traditional approach of using RNNs to model a student's historical interaction and instead makes predictions by taking into account relevant exercises from his past interactions.} SAKT has been shown to be far more efficient than the RNN-based KT model. 
    \item \textbf{EKT} \cite{liu2019ekt} is an extension to the Exercise-Enhanced Recurrent Neural Network (EERNN) framework. Compared to EERNN, EKT further introduces information about the knowledge concepts present in each exercise.
    \item \textbf{SAINT+} \cite{choi2020towards} , the first Transformer-based Knowledge Tracing model, is unique in that it introduces exercise information as well as student response information separately, while at the same time it embeds two temporal features, elapsed time and lag time, into the embedding of student response information.

\end{itemize}

\subsection{Experiment discussion}



Table \ref{table2} compares the predictive performance of Bi-CLKT and its variants with other mainstream baseline methods for ML and DL. The \emph{Area Under the curve (AUC)} and \emph{Accuracy (ACC)} are used as evaluation metrics.

\advice{The empirical performance is summarised in Table \ref{table2}. Overall, we can find that our proposed model shows competitive performance on all datasets. Bi-CLKT consistently outperforms all other baseline KT models by a wide margin. The competitive performance validates the superior performance of our proposed contrastive learning model for the knowledge tracing task. While existing baselines have achieved sufficiently high performance, our approach Bi-CLKT still pushes this bound forward. Furthermore, we note that Bi-CLKT competes with models based on the latest deep learning methods on all four datasets.}





\subsection{Overall Performance}

Table ~\ref{table2} summarises the results of the AUC and ACC comparisons for all baseline methods on the four datasets. From the results, we observe that our Bi-CLKT model achieves the best performance on all four datasets, ASSISTment 2009, ASSISTment 2015, ASSISTment Chall and STATICS 2011, which validates the validity and superiority of our model. Specifically, our proposed Bi-CLKT model achieves at least $5\%$ improvement than the other baseline models. In the baseline models, deep learning models consistently perform better than traditional machine learning models like BKT. This justifies the current research trend towards deep learning methods. We can also see that DKVMN performs slightly worse than DKT on average, as building states for each concept may lose information about the relationships between concepts. Furthermore, SAKT performs worse than our model, suggesting that there is a difference between exploiting higher-order concept-exercise relationships by selecting the most relevant exercises and performing interactions. Finally we can see that SAINT+, the best performing of the baseline models, is the first model to apply the transform modelling framework to the Knowledge Tracing task, which reflects the good adaptation of transform learning to this task. To further dissect our model, we provide sufficient ablation studies on the internal constructs of the model in the following sections.

\begin{table*}[ht]

\caption{Predictive performance. The best performing runs per metric per dataset are marked in boldface}
\centering
\begin{tabular}{c|cccccccc}
\hline
\multirow{2}{*}{Augmentation} & \multicolumn{2}{c}{ASSISTment 2009} & \multicolumn{2}{c}{ASSISTment 2015} & \multicolumn{2}{c}{ASSISTment Chall} & \multicolumn{2}{c}{STATICS 2011} \\ \cline{2-9} 
                              & \textit{AUC}      & \textit{ACC}     & \textit{AUC}     & \textit{ACC}     & \textit{AUC}      & \textit{ACC}     & \textit{AUC}    & \textit{ACC}   \\ \hline
Uniform                       & 0.864             & 0.786            & 0.748            & 0.749            & 0.77              & 0.753            & 0.852           & 0.815          \\
Degree                        & 0.869             & 0.795            & \textbf{0.765}   & \textbf{0.757}   & 0.773             & \textbf{0.764}   & 0.858           & 0.821          \\
PageRank                      & \textbf{0.875}    & \textbf{0.802}   & 0.757            & 0.752            & \textbf{0.775}    & \textbf{0.764}           & \textbf{0.865}  & \textbf{0.835} \\ \hline
\end{tabular}

\label{table3}
\end{table*}

\begin{table*}[h]
\caption{The Effect of Embedding Propagation Layer. The best performing runs per metric per dataset are marked in boldface}
\centering
\begin{tabular}{cc|cccccccc}
\hline
\multirow{2}{*}{Variants}                                                          & \multirow{2}{*}{Embedding} & \multicolumn{2}{c}{ASSISTment 2009} & \multicolumn{2}{c}{ASSISTment 2015} & \multicolumn{2}{c}{ASSISTment Chall} & \multicolumn{2}{c}{STATICS 2011} \\ \cline{3-10} 
                                                                                   &                            & \textit{AUC}      & \textit{ACC}     & \textit{AUC}     & \textit{ACC}     & \textit{AUC}      & \textit{ACC}     & \textit{AUC}    & \textit{ACC}   \\ \hline
\multirow{3}{*}{\textbf{\begin{tabular}[c]{@{}c@{}}Bi-CLKT-M\end{tabular}}} & C2C                        & 0.838             & 0.768            & 0.762            & 0.746            & 0.733             & 0.752            & 0.802           & 0.777          \\
                                                                                   & E2E                        & 0.83              & 0.762            & 0.748            & 0.748            & 0.744             & 0.747            & 0.857           & 0.788          \\
                                                                                   & Concate                    & 0.862             & 0.795            & 0.764            & 0.752            & 0.769             & 0.751            & 0.859           & 0.833          \\ \hline
\multirow{3}{*}{\textbf{\begin{tabular}[c]{@{}c@{}}Bi-CLKT-R\end{tabular}}}   & C2C                        & 0.847             & 0.784            & 0.764            & 0.754            & 0.761             & 0.76             & 0.849           & 0.817          \\
                                                                                   & E2E                        & 0.859             & 0.795            & 0.765            & 0.755            & 0.761             & 0.761            & 0.864           & 0.828          \\
                                                                                   & Concate                    & \textbf{0.875}    & \textbf{0.802}   & \textbf{0.765}   & \textbf{0.757}   & \textbf{0.775}    & \textbf{0.764}   & \textbf{0.865}  & \textbf{0.835} \\ \hline
\end{tabular}
\label{table4}
\end{table*}

\subsection{Ablation Studies}
To get insights into the effect of each module in Bi-CLKT, we design several ablation studies. Specifically, we further investigate the effectiveness of three important components of our proposed model: (1) augmentation methods; (2) dmbedding methods; (3) the predictive layer. We set a total of nine comparative settings and report the performances in Table \ref{table3} and Table \ref{table4}.

\subsection{Effects of Augmentation methods}
We observed that all three variants of Bi-CLKT with different node centrality measures outperformed the existing KT baseline model on all datasets. We also note that the augmentation with degree and PageRank centrality are two powerful variants that achieve the best or competitive performance on all datasets. \advice{Specifically, the augmented variant with PageRank centrality works best on the ASSISTment 2009, ASSISTment Chall and STATICS 2011 datasets, while only on ASSISTment 2015 does the augmented variant with Degree centrality outperform PageRank. This shows that our final model has good generalization and is not limited to a specific choice of augmentation method for different datasets.} 

\subsection{Effects of Different Embedding Methods}
Since the two implicit relations "exercise-to-exercise" (E2E) and "concept-to-concept" (C2C) are constructed separately in our Bi-CLKT model for the subgraphs. To better verify the role of these two embeddings in the model, we adopted the ``exercise-to-exercise" (E2E) and ``concept-to-exercise" (C2C) subgraphs separately as the attribute of each exercise and compared with their concatenation. From Table \ref{table4}, we can see that the results obtained by concatenation are significantly better than those obtained by using either embedding alone. Moreover, due to the difference in the prediction layer mechanism, C2C embedding is overall better than E2E embedding in this variant of Bi-CLKT-M, especially on the ASSISTment 2009 and ASSISTment 2015 datasets. In contrast, in the DKT variant, the use of E2E embedding alone gives better results, specifically on the ASSISTment 2009 and STATICS 2011 datasets. In general, both embeddings work well separately, and the best results are obtained when they are combined.

\subsection{Effects of Different Predictive Layers}
To improve the performance of the models, we used two distinct prediction mechanisms, Bi-CLKT-R and Bi-CLKT-M, which apply Recurrent Neural Network and Memory-augmented Neural Networks, respectively, in the prediction layer. In particular, on the ASSIST09 and STATICS 2011 datasets, our Bi-CLKT-R model achieves an AUC of over 0.85 and an ACC of over 0.8. Compared with Bi-CLKT-M, there are some slight differences with this variant of Bi-CLKT-R, despite the fact that this variant improves overall performance by at least 3\% over all other baseline KT models. We can find that in the ASSISTment 2015 dataset the two variants perform fairly close to each other, however, in the other three datasets there is a gap of at least 2\% between the two variants. Therefore, in the final model selection we chose Bi-CLKT-R as the predictive layer of the model for best results.


\section{Conclusion}
We transformed the traditional Knowledge Tracing problem into a graph form and proposed Bi-CLKT model that exploits contrastive learning to learn from large amounts of unlabelled data. Bi-CLKT consists of three main parts: subgraph establishing, contrastive learning and performance prediction. In the contrastive learning part, we adopt two different contrastive learning frameworks, local-local and global-global, for the ``exercise-to-exercise" (E2E) and  ``concept-to-concept" (C2C) implicit relationships respectively. The final ``exercise-to-exercise" (E2E) and ``concept-to-concept" (C2C) embeddings were obtained by node-level and graph-level GCN, and are concatenated together as attributes for each exercise into the prediction layer. Our proposed approach achieved significantly better performance compared to previous state-of-the-art methods for Knowledge Tracing tasks on multiple challenging datasets.


\bibliographystyle{elsarticle-num}
\bibliography{refs}




\end{document}